\documentclass[letterpaper, 10 pt, journal, twoside]{IEEEtran}

\usepackage{balance}
\usepackage{caption}
\usepackage{graphicx}
\usepackage{cite}
\usepackage{mathtools}
\usepackage{amsmath,amsfonts,amssymb}
\usepackage{booktabs}
\usepackage{array}
\usepackage{multirow}
\usepackage{makecell}
\usepackage{threeparttable}
\usepackage{hyperref}

\begin{document}
\title{FreeDOM: Online Dynamic Object Removal Framework for Static Map Construction Based on Conservative Free Space Estimation}

\author{Chen Li, Wanlei Li, Wenhao Liu, Yixiang Shu, and Yunjiang Lou
\thanks{Chen Li, Wanlei Li, Wenhao Liu, Yixiang Shu, and Yunjiang Lou are with the School of Intelligence Science and Engineering, College of Artificial Intelligence, Harbin Institute of Technology Shenzhen, Shenzhen 518055, China. E-mail: \{23S153029, 19b953034\}@stu.hit.edu.cn; liuwen810\_718@163.com; 755593654@qq.com; louyj@hit.edu.cn.}
}

\maketitle
\begin{abstract}
Online map construction is essential for autonomous robots to navigate in unknown environments. However, the presence of dynamic objects may introduce artifacts into the map, which can significantly degrade the performance of localization and path planning. To tackle this problem, a novel online dynamic object removal framework for static map construction based on conservative free space estimation (FreeDOM) is proposed, consisting of a scan-removal front-end and a map-refinement back-end. First, we propose a multi-resolution map structure for fast computation and effective map representation. In the scan-removal front-end, we employ raycast enhancement to improve free space estimation and segment the LiDAR scan based on the estimated free space. In the map-refinement back-end, we further eliminate residual dynamic objects in the map by leveraging incremental free space information. As experimentally verified on SemanticKITTI, HeLiMOS, and indoor datasets with various sensors, our proposed framework overcomes the limitations of visibility-based methods and outperforms state-of-the-art methods with an average F1-score improvement of 9.7\%. The source code is released to support future research in the area\footnote{Code: \url{https://github.com/LC-Robotics/FreeDOM}}. A demonstration video is available on YouTube\footnote{Video: \url{https://youtu.be/5w5p5S4NJ4E}}. 
\end{abstract}

\section{Introduction}
\begin{figure}[t]
    \centering
    \vspace{-0.2cm}
    \includegraphics[width=8.5cm]{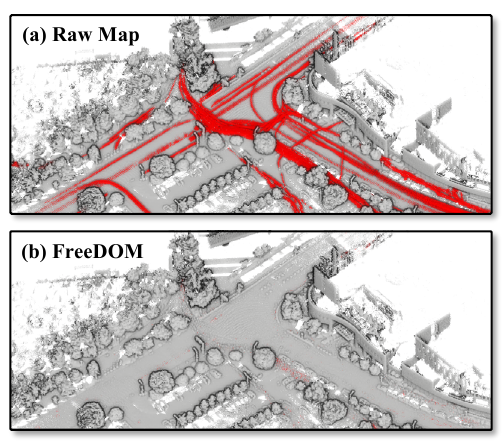}
    \vspace{-0.2cm}
    \caption{Comparison between (a) raw point cloud map and (b) map generated by FreeDOM (dynamic objects are highlighted in red).
}
    \label{fig1}
    \vspace{-0.6cm}
\end{figure}

\IEEEPARstart{O}{nline} construction of a clean static map is essential for localization, navigation, and exploration of autonomous robots in \textit{unknown} environments. Since autonomous robots are frequently deployed in \textit{dynamic} environments that contain various moving objects, such as pedestrians and vehicles, dynamic object removal (DOR) is inevitable to avoid the \textit{ghost trail effect} \cite{pomerleau2014long}, as shown in Fig. \ref{fig1} (a). Such trails may generate erroneous feature points during the SLAM process \cite{xu2022fast,vizzo2023kiss}, leading to localization drift or even failure \cite{9636624}. Furthermore, those trails may obstruct path planning, resulting in failures in navigation and exploration.

Dynamic object removal can be generally divided into two categories: a) offline DOR with all scans\cite{schauer2018peopleremover,kim2020remove,9568799,lim2021erasor,lim2023erasor2,10533672}, and b) online DOR at the time of each scan \cite{yoon2019mapless,milioto2019rangenet++,chen2021moving,mersch2022receding,mersch2023building,schmid2023dynablox,hornung2013octomap,10496850,fan2022dynamicfilter,yan2023rh}. Offline DOR typically requires long sequences of scans and cannot incrementally construct static maps, resulting in very long computation times when updating the map, and thus it is inapplicable for online robotics applications. 

Online DOR can be further categorized into scan-removal \cite{yoon2019mapless,milioto2019rangenet++,chen2021moving,mersch2022receding,mersch2023building,schmid2023dynablox} and map-removal \cite{hornung2013octomap,10496850,fan2022dynamicfilter,yan2023rh}. Online scan-removal generally relies solely on short sequences and can effectively remove most dynamic objects in the current scan. In recent years, deep learning-based methods \cite{milioto2019rangenet++,chen2021moving,mersch2022receding,mersch2023building} have demonstrated significant potential for scan-removal of dynamic objects. Chen et al. utilized residual images from previous scans to explore sequential information \cite{chen2021moving}. Mersch et al. proposed using a Sparse 4D CNN network to predict dynamic objects from continuous point cloud sequences \cite{mersch2022receding}. However, deep learning-based methods rely heavily on manually labeled data and often suffer from degraded performance when applied to unseen environments or different sensor setups. 

In contrast, non-learning-based methods do not rely on pre-trained models, therefore unaffected by this issue. Yoon et al. detected dynamic objects based on scan-to-scan differences \cite{yoon2019mapless}. Schmid et al. proposed integrating visibility information to estimate free space, detecting dynamic objects based on the estimated free space \cite{schmid2023dynablox}. However, directly integrating scan-removal results into static maps may lead to the accumulation of misclassified dynamic points \cite{mersch2023building}. To eliminate these errors, map-removal is necessary. 

Map-removal typically uses geometric or visibility cues to directly remove dynamic objects from the map. Among them, ground assumption-based methods \cite{9568799,lim2021erasor,lim2023erasor2,yan2023rh,10533672} presume that all dynamic objects are in contact with flat ground and detect potential dynamic regions based on scan-to-map height differences. However, autonomous robots often operate in unstructured scenarios, such as stairs, ramps, narrow tunnels, or naturally rugged terrain, where ground assumption-based methods frequently fail. 

Visibility-based methods utilize the visibility cue that if a query point is observed behind other points, those points are dynamic. Based on this cue, OctoMap \cite{hornung2013octomap} enumerates all voxels traversed by measurements using raycasting \cite{amanatides1987fast} and updates these voxels as \textit{free} to remove dynamic objects. However, due to \textit{incidence angle ambiguity} \cite{lim2021erasor}, raycasting-based methods typically generate a significant number of FPs (false positives, static points falsely classified as dynamic) on surfaces with large incidence angles (e.g., the ground). To address this, DUFOMap \cite{10496850} marks a voxel as \textit{free} only if all its neighboring voxels are traversed in a single scan, but this approach struggles with sparse sensors. Furthermore, raycasting-based methods are computationally expensive for online applications, particularly at high resolution. 

To reduce computational costs, Kim and Kim proposed an offline DOR method that uses range images instead of raycasting \cite{kim2020remove}. This method projects the map onto each frame and removes map points with ranges smaller than those of the query point. However, map-removal often rely on information after the arrival of dynamic objects, which may result in residual dynamic points in certain scenarios (e.g., trailing vehicles). To address this issue, Fan et al. accumulate \(n\) consecutive scans into a local submap and perform scan-to-map removal for each scan to utilize visibility information before the arrival of dynamic objects \cite{fan2022dynamicfilter}. However, this method significantly increases computational costs and can only utilize history visibility information within \(n\) scans. 

Instead of directly performing scan-to-map removal, Schmid et al. proposed integrating visibility information from each scan to estimate conservative free space \cite{schmid2023dynablox}. Their method removes dynamic objects based on the cue that when a point falls into space that is known to be free, the point must have moved there and thus be dynamic. By estimating conservative free space, their method avoids the FP issue of raycasting and effectively leverages all visibility information before the arrival of dynamic objects without significantly increasing computational costs. However, since their method focuses on real-time scan-removal, it does not fully exploit visibility information after the arrival of dynamic objects, leading to the accumulation of misclassified dynamic objects. 

In this paper, we propose a novel online DOR framework for static map construction based on conservative free space estimation, referred to as FreeDOM. To achieve quality maps with reduced computational costs, we propose a multi-resolution map structure that estimates free space at a lower resolution while representing the static map at a higher resolution. Additionally, we propose raycast enhancement to recover free space in directions without background. FreeDOM consists of a scan-removal front-end for fast dynamic object removal and a map-refinement back-end for continuous map refinement. The scan-removal front-end integrates visibility information from each scan into free space and segments the LiDAR scan. The map-refinement back-end further removes residual dynamic objects from the map by utilizing incremental free space information. As FreeDOM does not rely on any environmental assumptions or pre-trained models, it is robust to various dynamic object classes and environments. 

Main contributions of the paper are as follows:
\begin{itemize}

\item FreeDOM, a dynamic object removal framework for static map construction consisting of a scan-removal front-end and a map-refinement back-end. 

\item A multi-resolution map structure for efficient free space estimation, map integration, and map clearing.

\item A raycast enhancement method that recover free space information in directions without background. 

\item The proposed method is evaluated across diverse scenarios and sensors, achieving state-of-the-art performance.
\end{itemize}

\section{Problem Statement}
Let us consider online DOR for static map construction using continuous LiDAR scans. Given a LiDAR scan \({}_SP^{(t)}=\{{}_Sp^{(t)}_i\in \mathbb{R}^3\}\) in the sensor frame \(S\) of the current time step \(t\) and its corresponding pose \(^W_ST^{(t)} \in SE(3)\) in the world frame \(W\), the online static map construction process can be described as
\begin{equation}
    \hat{\mathcal{M}}^{(t)} = \nu\left(R\left(\hat{\mathcal{M}}^{(t-1)},\ \mathcal{H}^{(t-1)},\ {}_SP^{(t)},\  ^W_ST^{(t)}\right)\right)
\end{equation}
\begin{equation}
    \mathcal{H}^{(t)} = U\left(\hat{\mathcal{M}}^{(t-1)}, \mathcal{H}^{(t-1)},\ {}_SP^{(t)},\ {}^W_ST^{(t)}\right)
\end{equation}
where \(\hat{\mathcal{M}}^{(t)}\) is the updated static map, \(\mathcal{H}^{(t)}\) is the updated algorithm state, which can include historical information, \(\nu(\cdot)\) denotes voxelization. 
The functions \(R\) and \(U\) incrementally update the static map and the algorithm state, respectively, using the current inputs \({}_SP^{(t)}\) and \(^W_ST^{(t)}\). The goal is to construct a complete and clean static map \(\hat{\mathcal{M}}^{(t)}\), i.e. 
\begin{equation}
    \displaystyle \max_{\hat{\mathcal{M}}^{(t)}} | \hat{\mathcal{M}}^{(t)} \cap \mathcal{M}^{(t)}_{sta} | \ \text{and} \ \displaystyle \min_{\hat{\mathcal{M}}^{(t)}} | \hat{\mathcal{M}}^{(t)} \cap \mathcal{M}^{(t)}_{dyn} |
\end{equation}
where \(| \cdot |\) counts the number of voxels, \(\mathcal{M}^{(t)}_{sta}\) and \(\mathcal{M}^{(t)}_{dyn}\) are the ground truth static map and dynamic map, defined as:
\begin{equation}\scalebox{1.0}{$
    \mathcal{M}^{(t)}_{sta} = \nu\left( \bigcup^{t}_{\tau =0} \ {}^W_ST^{(\tau)} {}_SP^{(\tau)}_{sta} \right)
$}\end{equation}
\begin{equation}\scalebox{1.0}{$
    \mathcal{M}^{(t)}_{dyn} = \nu\left( \bigcup^{t}_{\tau =0} \ {}^W_ST^{(\tau)} {}_SP^{(\tau)} \right) \setminus \mathcal{M}^{(t)}_{sta}\text{.}
$}\end{equation}
Here \({}_SP^{(\tau)}_{sta}\) is the static part of scan \({}_SP^{(\tau)}\). Intuitively, the goal is to maximize the preservation of static points while minimizing the preservation of dynamic points. For online robotics applications, the processing time of \(R\) and \(U\) should not exceed the time between time steps \(t\) and \(t+1\). 

\begin{figure*}[t]
    \centering
    %\vspace{-0.2cm}
    \includegraphics[width=17.5cm]{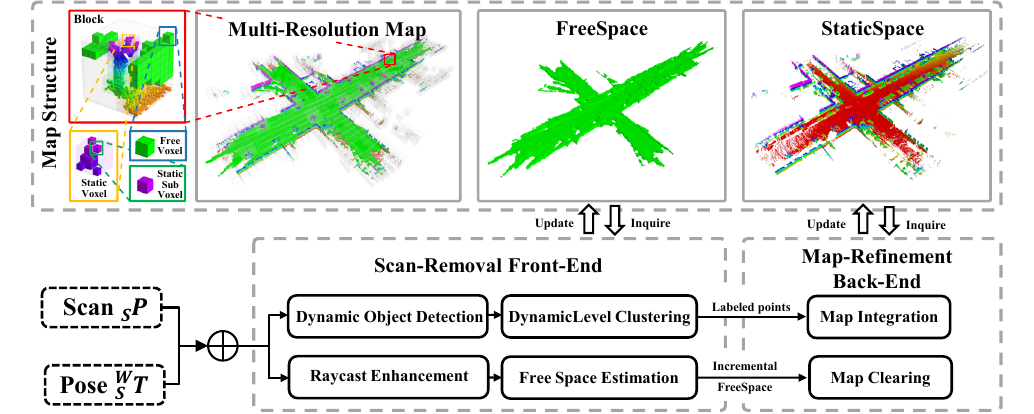}
    \vspace{-0.1cm}
    \caption{Overview of FreeDOM: the multi-resolution map, representing \textit{FreeSpace} (green voxels) at voxel level and \textit{StaticSpace} (rainbow-colored voxels) at subvoxel level, is shown at the top of Fig. \ref{framework}. Given the LiDAR scan \({}_SP\) and estimated pose \({}^W_ST\), the \textit{scan-removal front-end} estimates \textit{FreeSpace} and segments the LiDAR scan. The \textit{map-refinement back-end} further removes residual dynamic points utilizing incremental \textit{FreeSpace}.
}
    \label{framework}
    \vspace{-0.6cm}
\end{figure*}

\section{Implementation}
A point must be dynamic when it falls into free space. Based on this spatial motion cue, FreeDOM removes moving objects in LiDAR scans using the estimated free space and refines the static map by removing residual dynamic points as free space estimation improves.
\subsection{Multi-Resolution Map Structure}
In this section, we introduce the multi-resolution map structure of FreeDOM, as shown in Fig. \ref{framework}. FreeDOM includes \textit{FreeSpace}, which represents free space information, and \textit{StaticSpace}, which represents occupancy information. To achieve quality maps with reduced computational cost, we use a multi-resolution map structure to represent \textit{FreeSpace} and \textit{StaticSpace}, estimating free space at a lower resolution and constructing the static map at a higher resolution.

\textbf{Multi-resolution:}
In \textit{FreeSpace}, we evenly divide the global space into blocks of size \(s_b\), and each block is divided into voxels of size \(s_v\). In \textit{StaticSpace}, each voxel is further subdivided into subvoxels of size \(s_s\), which defines the resolution of the static map. We define the depths of subvoxels, voxels, and blocks as \(d_s\), \(d_v\) and \(d_b\), respectively. The relationship between their sizes is given by
\begin{equation}
    s_j = 2^{(d_j - d_i)} s_i, \quad i,j = s,v,b \ \text{.}
\end{equation}
In our experiments, we use a subvoxel size of \(s_s = 0.1 \text{m}\) for outdoor environments and \(0.05 \text{m}\) for indoor environments. The voxel size \(s_v\) directly impacts the conservativeness of free space, which we explore further in Section IV (Impact of Conservativeness). The block size \(s_b\), although independent of DOR performance, is empirically set to \(s_b = 3.2 \text{m}\) to balance computational efficiency and memory consumption.

To incrementally map potentially unbounded environments, blocks are dynamically allocated and indexed using 3D hash function. Voxels and subvoxels are represented in dense grids to enhance access speed. In order to obtain the subvoxel, voxel, or block where a point \(p\) is located, we calculate the global and local indices as follows:
\begin{equation}
    I_{i} = \left \lfloor \frac{1}{s_{i}} \left [ p_x, p_y, p_z \right ]\right \rfloor , \quad i = s,v,b
\end{equation}
\begin{equation}
    I_{j,local} = I_{j} \& m_{j} , \quad j = s,v
\end{equation}
where \(I_i\) is the global index of point \(p\) at resolution \(i\), \(I_{j,local}\) is the local index of \(I_{j}\), \(\left [ p_x, p_y, p_z \right ]\) are coordinates of point \(p\) in the world frame, and \(I_{j} \& m_{j}\) denotes a coordinate-wise bitwise \texttt{AND} between \(I_{j}\) and the binary mask \(m_{j}\) of the resolution \(j\) (i.e., \(d_s = 0, d_v = 2 \ \text{and} \ d_b = 5 \text{, then} \ m_s = 00000011, m_v = 00000111\)). The index at higher resolution can be converted to the index at lower resolution through coordinate-wise bitwise right shift operations \(\gg\): 
\begin{equation}
I_j = I_i \gg (d_j - d_i), \quad d_j > d_i, \quad i,j = s,v,b \ \text{.}
\end{equation}

\textbf{FreeSpace:} \textit{FreeSpace} represents free space information at voxel resolution. In \textit{FreeSpace}, voxels are referred to as \textit{FreeVoxel}s. When all \textit{FreeVoxel}s in a block are \textit{free}, they are released to save storage space, and the block is marked as \textit{free}. Each \textit{FreeVoxel} represents whether the voxel is \textit{free} \(f\) and two variables related to conservative free space estimation:
\begin{equation}
    \left\{ f,n_{f},n_{o}\right\}
\end{equation}
where \(n_{f}\) and \(n_{o}\) denote the number of consecutive observations as \textit{free} and \textit{occupied}, respectively.

\textbf{StaticSpace:} \textit{StaticSpace} represents occupancy information at subvoxel resolution. In \textit{StaticSpace}, voxels are referred to as \textit{StaticVoxel}s, and subvoxels as \textit{StaticSubVoxel}s. Each \textit{StaticSubVoxel} represents the time step \(t_o\) and the DynamicLevel \(d\) of the point:
\begin{equation}
    \left\{ t_o, d \right\} \text{.}
\end{equation}
Based on the degree of dynamism, points are classified into DynamicLevels: \{\textit{static}, \textit{aggressive}, \textit{moderate}, \textit{conservative}\}, abbreviated and totally ordered as \{\textit{s} \(\prec\) \textit{a} \(\prec\) \textit{m} \(\prec\) \textit{c}\}. 

\subsection{Scan-Removal Front-End}
The scan-removal front-end uses the raycasting method to integrate visibility information from each scan into \textit{FreeSpace} and segments the LiDAR scan in real time based on the estimated \textit{FreeSpace}. 

\begin{figure}[t]
    \centering
    \includegraphics[width=8.5cm]{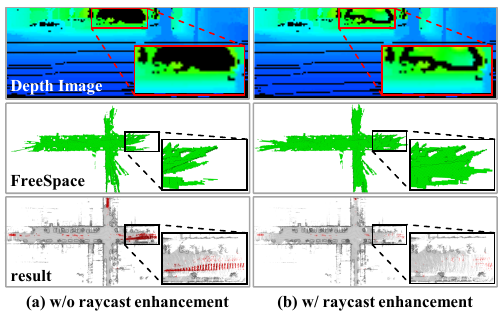}
    \vspace{-0.1cm}
    \caption{Effect of \textit{Raycast Enhancement} on SemanticKITTI 07: \textit{Raycast Enhancement} recovers free space information in directions without background, resulting in better free space estimation (green voxels) and fewer residual dynamic points (red points). 
}
    \label{raycast_enh}
    \vspace{-0.6cm}
\end{figure}

\textbf{Raycast enhancement:} 
Visibility-based methods are based on the cue that if a query point is observed behind other points, those points are dynamic. However, due to the sensor characteristics of LiDAR, points may not be returned in directions that are too far away or have no background (e.g., sky). We propose an effective strategy to recover free space information in these directions, as shown in Fig. \ref{raycast_enh}. First, we map the point cloud \({}_SP\) to the depth image \(I = \left( I_{i,j} \right) \in \mathbb{R}^{m \times n}\):
\begin{equation}
    I_{ij} = \displaystyle \min_{p\in {}_SP_{ij}}r(p)
\end{equation}
where \(r(p)\) represents the range of point \(p\) within the sensor frame, \({}_SP_{ij}\) is a subset of \({}_SP\) defined as:
\begin{equation}
    {}_SP_{ij} = \big\{ p \in {}_SP \mid \lfloor \frac{\phi(p)-\phi_{min}}{\phi_{res}} \rfloor = i, \lfloor \frac{\theta(p)-\theta_{min}}{\theta_{res}} \rfloor = j \big\} \text{.}
\end{equation}
Here, \(\phi(p) \in \left[ \phi_{min}, \phi_{max} \right]\) and \(\theta(p) \in \left[ \theta_{min}, \theta_{max} \right]\) are the azimuth and elevation angles of point \(p\) in the sensor frame, while \(\phi_{res}\) and \(\theta_{res}\) are the angular resolutions of the depth image. The number of pixels \(m\) and \(n\) is determined by the FOV and point density of the LiDAR. The region without LiDAR measurement \(R\) is used as the region for raycast enhancement:
\begin{equation}
    R = \{ (i, j) \in M_{ij} \mid {}_SP_{ij} = \varnothing\}
\end{equation}
where \(M_{ij}\) is a FoV mask determined by the LiDAR scanning pattern and fixed occlusions. However, the absence of point clouds in a direction might also be due to low-reflectivity. Therefore, we set a maximum raycast enhancement range \(r_{\text{max}}\), within which LiDAR can detect most objects (i.e., the detection distance of Velodyne HDL-64 at 10\% reflectivity is \(50 \text{m}\), so \(r_{\text{max}} = 50 \text{m}\)). Since the depth of low-reflectivity objects within \(r_{\text{max}}\) is typically related to their surroundings (e.g., tires), we use weighted averaging to estimate the possible depth of these objects within the raycast enhancement region \(R\) and introduce a safety margin \(r_m\) to avoid false positives:
\begin{equation}
    P = \mathcal{P} \left ( I , R \right )
\end{equation}
\begin{equation}
E_{ij} = \left\{\begin{matrix}
\max\{\min\{ P_{ij} - r_{m}, r_{\text{max}} \}, 0\} & if \left( i,j \right) \in R \\
0 & \text{otherwise}
\end{matrix}\right.
\end{equation}
where \(\mathcal{P} \left ( I , R \right )\) represents the weighted average filling of the image \(I\) within the region \( R \). \(E\) is the recovered depth of free space. We map \(E\) to a point cloud \({}_SP_E\) to represent the recovered free space boundary and convert it to \({}_WP_E\).

\begin{figure}[t]
    \centering
    \vspace{-0.2cm}
    \includegraphics[width=8.5cm]{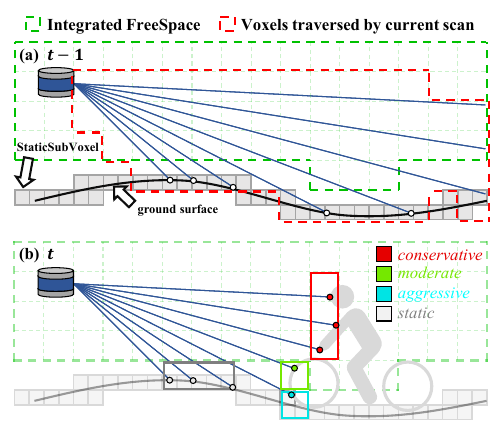}
    \vspace{-0.4cm}
    \caption{Procedure of the \textit{scan-removeal front-end}. (a) \textit{Free space estimation}: integrates visibility information from each scan into \textit{FreeSpace} using a conservative strategy. (b) \textit{Real-time dynamic object removal}: assigns DynamicLevel labels to points based on the estimated \textit{FreeSpace}. 
}
    \label{raycast}
    \vspace{-0.6cm}
\end{figure}

\begin{figure*}[t]
    \centering
    %\vspace{-0.2cm}
    \includegraphics[width=17.5cm]{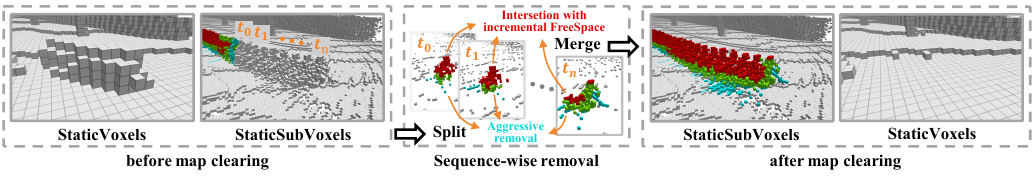}
    \vspace{-0.2cm}
    \caption{Procedure of \textit{map clearing}: Sequence-wise removal is performed utilizing incremental \textit{FreeSpace}, resulting in a clean and complete static map. 
}
    \label{clearing}
    \vspace{-0.6cm}
\end{figure*}

\textbf{Free space estimation:} 
To estimate the free space, we obtain all \textit{FreeVoxel}s \( V_o\) occupied by the scan, and use raycasting to obtain all \textit{FreeVoxel}s \( V_f\) traversed by the sensor measurement \({}_WP\) and the recovered free space boundary \({}_WP_E\). To accelerate the raycasting process, we skip all blocks already marked as \textit{free}. Considering sensor noise, pose uncertainty, and the \textit{incidence angle ambiguity}, we adopt a spatial and temporal conservative strategy, as shown in Fig. \ref{raycast}. A voxel is only considered free when all 
voxels in its neighborhood are consistently traversed. For each \textit{FreeVoxel} \(v \in V_f\) with \(f(v) = 0\), update the state of \(v\):
\begin{equation}
    n_f^{(t)} = n_f^{(t-1)} + 1
\end{equation}
\begin{equation}
    n_o^{(t)} = 0
\end{equation}
\begin{equation}
    f^{(t)} = \left\{\begin{matrix}
    1 & \text{if} \ \forall v' \in N_m \left( v \right) , \ n_f^{(t)}\left( v' \right) \geq \tau_f \\
    0 & \text{otherwise}
    \end{matrix}\right.
\end{equation}
where \({\cdot}^{(t)}\) represents the value of \(\cdot\) in the current time step \(t\), while \({\cdot}^{(t-1)}\) represents the value of \(\cdot\) in the previous time step \(t-1\). \(\tau_f\) is the temporal conservativeness duration, which will be discussed in Section IV (Impact of Conservativeness). \(N_m \left( v \right)\) is the neighborhood of voxel \(v\), including \(v\) itself. To reduce computational cost, we apply a small neighborhood (27-neighborhood in our experiment) and regulate the spatial conservativeness by adjusting the voxel size \(s_v\). 

However, in cases of significant drift in state estimation, static objects may enter the \textit{FreeSpace} and cause a large number of false positives. Therefore, we allow \textit{FreeSpace} to recover after being continuously occupied. For each \textit{FreeVoxel} \(v \in V_o\), update the state of \(v\):
\begin{equation}
    n_o^{(t)} = n_o^{(t-1)} + 1
\end{equation}
\begin{equation}
    n_f^{(t)} = 0
\end{equation}
\begin{equation}
    f^{(t)} = \left\{\begin{matrix}
    0 & \text{if} \ n_o^{(t)}\left ( v \right ) \geq \tau_r \\
    1 & \text{otherwise}
    \end{matrix}\right.
\end{equation}
where \(\tau_r\) is the recovery threshold of \textit{FreeSpace} when it is continuously occupied. Notably, when \(v\) is not in \textit{FreeSpace} and \(n_o^{(t)}\left ( v \right ) \geq \tau_r\), we also set \(f(v')=0\) for all \(v' \in N_m(v)\) to ensure the spatial conservativeness of \textit{FreeSpace}. However, slowly moving dynamic objects may also be misclassified as static ones. Therefore, we set a relatively large threshold (20 in our experiment) to prevent misclassification. 

\textbf{Real-time dynamic object removal:} 
A point must be dynamic when it falls into free space. Based on the estimated \textit{FreeSpace}, we can now detect dynamic objects using this spatial motion cue. Considering that the estimated \textit{FreeSpace} may deviate from the actual free space due to spatial conservativeness, insufficient observations, sensor noise, or pose estimation drift, we calculate the DynamicLevel of all scan voxels \(V_{scan}\), i.e., voxels occupied by the LiDAR scan \({}_WP\), based on their distance to \textit{FreeSpace} and the spatial continuity of dynamic objects, as shown in Fig. \ref{raycast}. First, we mark all scan voxels that fall into \textit{FreeSpace} as \textit{conservative}:
\begin{equation}
    V_{\textit{c}} = \{ v \in V_{scan} \mid f(v) = 1 \} \text{.}
\end{equation}
Then, we use the \textit{conservative} scan voxels \(V_c\) as the seed to grow \textit{moderate} scan voxels \(V_{m}\) and \textit{aggressive} scan voxels \(V_{a}\), the remaining scan voxels are categorized as \textit{static}:
\begin{equation}
    V_{\textit{m}} = \{ v \in (V_{scan} \setminus V_{\textit{c}}) \mid \exists v' \in N_{\textit{m}}(v), v' \in V_{\textit{c}}  \}
\end{equation}
\begin{equation}
    V_{\textit{a}} = \{ v \in (V_{scan} \setminus V_{\textit{c},\textit{m}}) \mid \exists v' \in N_{\textit{a}}(v), v' \in V_{\textit{m}}  \}
\end{equation}
\begin{equation}
    V_{\textit{s}} = V_{scan} \setminus V_{\textit{c},\textit{m},\textit{a}}
\end{equation}
where \(V_{\textit{c},\textit{m}} = V_{\textit{c}} \cup V_{\textit{m}}\), \(V_{\textit{c},\textit{m},\textit{a}} = V_{\textit{c}} \cup V_{\textit{m}} \cup V_{\textit{a}}\), \(N_{\textit{m}}(v)\) and \(N_{\textit{a}}(v)\) represent the neighborhoods of \(v\). To revert spatial conservativeness, we employ the same neighborhood \(N_{\textit{m}}(v)\) as in free space estimation. We futher adopt a neighborhood \(N_{\textit{a}}(v)\) (125-neighborhood in our experiment) for aggressive removal around detected dynamic objects. Finally, we assign the DynamicLevel to each point \(p \in {}_WP\) based on the DynamicLevel of the scan voxel they belong to. 

\begin{table*}[t]
\begin{center}
\caption{Quantitative comparison on KITTI, HeLiMOS (Ouster LiDAR), and indoor datasets. Methods with \textsuperscript{\dag} operate offline. }
\label{table:1}
\tabcolsep=0.9mm
\begin{tabular}{m{2.0cm}<{\raggedright} m{0.8cm}<{\centering} m{0.8cm}<{\centering} m{0.8cm}<{\centering} m{0.1cm}
                                        m{0.8cm}<{\centering} m{0.8cm}<{\centering} m{0.8cm}<{\centering} m{0.1cm}
                                        m{0.8cm}<{\centering} m{0.8cm}<{\centering} m{0.8cm}<{\centering} m{0.1cm}
                                        m{0.8cm}<{\centering} m{0.8cm}<{\centering} m{0.8cm}<{\centering} m{0.1cm}
                                        m{0.8cm}<{\centering} m{0.8cm}<{\centering} m{0.8cm}<{\centering}}
\toprule
  & \multicolumn{3}{c}{KITTI seq. 02} & & 
    \multicolumn{3}{c}{KITTI seq. 07} & & 
    \multicolumn{3}{c}{HeLiMOS 8292-8901}   & & 
    \multicolumn{3}{c}{Corridor}   & & 
    \multicolumn{3}{c}{Stairs} \\ 
\midrule
Method & PR & RR & F$_1$ & & 
         PR & RR & F$_1$ & & 
         PR & RR & F$_1$ & & 
         PR & RR & F$_1$ & & 
         PR & RR & F$_1$\\
\midrule
\makecell[l]{OctoMap}                           &79.71             &\textbf{99.97}    &88.70          &
                                                &79.70             &88.73             &83.97          &
                                                &83.07             &94.11             &88.24          &
                                                &88.04             &97.84             &92.69          &
                                                &89.59             &95.18             &92.30          \\
\makecell[l]{DUFOMap\textsuperscript{\dag}}     &96.68             &99.84             &98.23          &
                                                &95.02             &80.86             &87.37          &
                                                &92.41             &91.62             &92.01          &
                                                &97.59             &96.25             &96.92          &
                                                &96.03             &94.60             &95.31          \\
\makecell[l]{Removert\textsuperscript{\dag}}    &92.30             &98.64             &95.37          &
                                                &91.17             &57.79             &70.74          &
                                                &83.19             &90.14             &86.53          &
                                                &91.43             &92.67             &92.05          &
                                                &92.27             &\textbf{95.53}    &93.87          \\
\makecell[l]{ERASOR\textsuperscript{\dag}}      &95.52             &99.78             &97.60          &
                                                &91.87             &\textbf{98.74}    &95.18          &
                                                &91.42             &91.58             &91.50          &
                                                &87.05             &90.10             &88.55          &
                                                &63.79             &76.67             &69.64          \\
\makecell[l]{BeautyMap\textsuperscript{\dag}}   &92.93             &99.33             &96.02          & 
                                                &96.22             &84.58             &90.02          & 
                                                &94.27             &88.86             &91.49          & 
                                                &77.60             &83.46             &80.42          & 
                                                &54.27             &21.63             &30.93          \\
\makecell[l]{FreeDOM (Ours)}                    &\textbf{99.50}    &99.69             &\textbf{99.59} &
                                                &\textbf{98.76}    &97.90             &\textbf{98.33} &
                                                &\textbf{98.01}    &\textbf{95.74}    &\textbf{96.86} &
                                                &\textbf{99.59}    &\textbf{98.57}    &\textbf{99.08} &
                                                &\textbf{99.48}    &95.11             &\textbf{97.25} \\
\bottomrule
\end{tabular}
\end{center}
\vspace{-0.8cm}
\end{table*}

\subsection{Map-Refinement Back-End}
Directly integrating scan-removal results into static maps may lead to the accumulation of classification errors. To tackle this, we propose a map-refinement back-end to eliminate these  errors based on incremental information from free space estimation, as shown in Fig. \ref{clearing}.

\textbf{Map integration:} 
Due to the conservativeness of \textit{FreeSpace}, directly removing the regions of static map that intersect with incremental \textit{FreeSpace} would fail to remove dynamic objects that are close to static objects. To address this issue, we leverage the spatial association of historical scans to eliminate residual dynamic points without compromising the static part. Specifically, we store the time step \(t_o\) and the corresponding DynamicLevel \(d\) of the point that led to the occupancy. When integrating the static map, we update the status of the corresponding \textit{StaticSubVoxel} \(s\) for each point \(p \in {}_WP\) labeled by the front-end:
\begin{equation}
    d^{(t)}(s), t_o^{(t)}(s) = 
    \left\{
    \begin{matrix}
    d(p), t & \text{if } d(p) \prec d^{(t-1)}(s) \\
    d^{(t-1)}(s), t_o^{(t-1)}(s) &\text{otherwise.}
    \end{matrix}
    \right.
\end{equation}
To preserve more important low DynamicLevel information, we replace the DynamicLevel \(d\) and time step \(t\) of the subvoxel when the DynamicLevel of the new point \(p\) is lower than that of the subvoxel (i.e., \textit{aggressive} to \textit{static}). 

\textbf{Map clearing:} 
Even in highly dynamic environments, observations of static objects are often redundant. Therefore, we exploit the spatial association of historical scans to aggressively remove points around detected dynamic parts, as shown in Fig. \ref{clearing}. We first extract the incremental \textit{FreeSpace} \(V_i = \{ v \mid f^{(t)}(v) = 1 \land f^{(t-1)}(v) = 0 \}\) and all time steps \(T_q\) of \textit{StaticSubVoxels} that intersect with \(V_i\):
\begin{equation}
    T_q = \{ t_o(s) \mid v(s) \in V_i\}
\end{equation}
where \(v(s)\) denotes the voxel that contains subvoxel \(s\). For each time step \(t_q \in T_q\) , obtain all \textit{StaticVoxels} containing \textit{StaticSubVoxels} with \(t_o(s) = t_q\):
\begin{equation}
	V_{t_q} = \{ v(s) \mid t_o(s) = t_q \} \text{.}
\end{equation}
Similar to the front-end, we use these \textit{StaticVoxel}s as the seed to grow the update range \(V_{t_q}\):
\begin{equation}
    V_{t_q,\textit{c}} = V_i \cap V_{t_q}
\end{equation}
\begin{equation}
    V_{t_q,\textit{m}} = \{ v \in V_{t_q} \setminus V_{t_q,\textit{c}} \mid \exists v' \in N_{\textit{m}}(v) , v' \in V_{t_q,\textit{c}} \}
\end{equation}
\begin{equation}
    V_{t_q,\textit{a}} = \{ v \in V_{t_q} \setminus V_{t_q,\textit{c},\textit{m}} \mid \exists v' \in N_{\textit{a}}(v) , v' \in V_{t_q,\textit{m}} \} \text{.}
\end{equation}
Then we update the DynamicLevel of all \textit{StaticSubVoxel} \(s\) such that \( v(s) \in V_{t_o(s),i},\ i = \textit{c},\textit{m},\textit{a}\):
\begin{equation}
	d^{(t)}(s) =
	\left\{
    \begin{matrix}
	i & \text{if } d^{(t-1)}(s) \prec i \\
	d^{(t-1)}(s) & \text{otherwise.}
	\end{matrix}
    \right.
\end{equation}
Finally, we take the part of \textit{StaticSpace} where the DynamicLevel is \textit{static} as the final static map output:
\begin{equation}
    \hat{\mathcal{M}}^{(t)} = \{ s \mid d(s) = \textit{static} \} \text{.}
\end{equation}

\begin{figure}[t]
    \centering
    %\vspace{0.1cm}
    \includegraphics[width=8.5cm]{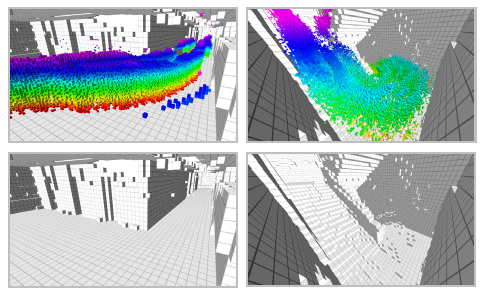}
    \vspace{-0.2cm}
    \caption{Indoor datasets. Top row: raw maps of Corridor (left) and Stairs (right), where the colored points represent dynamic objects detected by FreeDOM. Bottom row: corresponding static maps generated by FreeDOM. 
}
    \label{real}
    \vspace{-0.6cm}
\end{figure}

\section{Experiments and Evaluation}
\subsection{Experimental Setup}
\textbf{Datasets:}
To evaluate the performance of our method, we conduct quantitative comparisons against other state-of-the-art methods on the SemanticKITTI dataset \cite{behley2019semantickitti,geiger2012we}. In addition, due to the limited dynamic scenes and the single LiDAR type (Velodyne HDL-64) in SemanticKITTI, we further perform comparisons on the more dynamic Helimos dataset \cite{lim2024helimos}, which is collected using four different LiDARs. 

While both SemanticKITTI and HeLiMOS primarily focus on outdoor flat environments, we further collect an indoor dataset to evaluate the robustness of our method under varied environments. The indoor dataset consists of two sequences: one collected in a corridor using a Robosense Helios-32 LiDAR and the other in a staircase using a Livox mid-360 LiDAR, as shown in Fig. \ref{real}.  

\textbf{Metrics:}
To evaluate the quality of the constructed static map, we adopt the preservation rate (PR) and rejection rate (RR) as evaluation metrics \cite{lim2021erasor}, defined as follows:
\begin{equation}
    PR = \frac{| \nu(\hat{\mathcal{M}}) \cap \mathcal{M}_{sta} |}{| \mathcal{M}_{sta} |} \quad RR = 1 - \frac{| \nu(\hat{\mathcal{M}}) \cap \mathcal{M}_{dyn} |}{| \mathcal{M}_{dyn} |}
\end{equation}
here we generate the ground truth \(\mathcal{M}_{sta}\) and \(\mathcal{M}_{dyn}\) with a voxel size of \(0.2 \text{m}\) for outdoor datasets and \(0.1 \text{m}\) for indoor datasets, and compute PR and RR voxel-wise. To assess the overall performance, we also compute the F1 score. Since different methods may use different resolutions for map construction, we voxelize all generated maps using the same voxel size as the ground truth before evaluation.

\begin{figure}[t]
    \centering
    %\vspace{-0.2cm}
    \includegraphics[width=8.5cm]{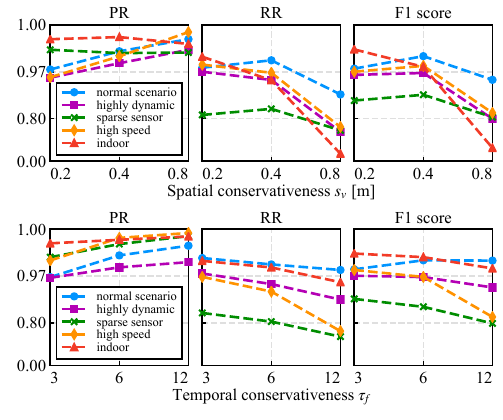}
    \vspace{-0.2cm}
    \caption{Performance impact of different spatial and temporal conservativeness levels on specified sequences.
}
    \label{params}
    \vspace{-0.6cm}
\end{figure}

\subsection{Evaluation}
\textbf{Impact of conservativeness:}
To investigate the impact of conservativeness under different conditions, we specified five sequences: HeLiMOS Ouster 6593-6758 (normal scenario), HeLiMOS Ouster 8292-8901 (highly dynamic), HeLiMOS Velodyne 11068-11662 (sparse sensor), SemanticKITTI 01 (high speed), and Corridor (indoor). As shown in Fig. \ref{params}, increasing conservativeness generally results in higher PR but lower RR. Notably, the optimal \(s_v\) that maximizes the F1 score is primarily influenced by the scale of the environment. In indoor settings, lower sensor noise and localization drift allow a smaller \(s_v\), which increases RR without compromising the static map. Based on this observation, we set \(s_v = 0.4\) for outdoor environments and \(s_v = 0.2\) for indoor environments. In contrast, the optimal \(\tau_f\) is affected by sensor density and platform speed. In sparse sensor setups and high-speed scenarios, limited environmental observations necessitate more aggressive and faster free space estimation. Accordingly, we set \(\tau_f = 3\) for sparse sensors and high-speed scenarios, and \(\tau_f = 6\) for other settings. 

\begin{figure*}[t]
%\vspace{-0.4cm}
\centering
\begin{minipage}[t]{0.19\linewidth}
\centering
\includegraphics[width=3.4cm,height=7.8cm]{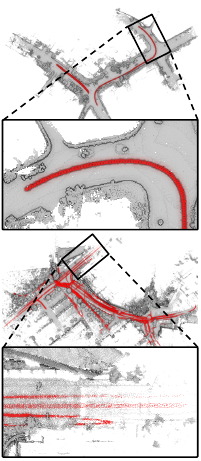}
%\subcaption{Original map}
\centerline{\small (a) Raw map}
%\caption{original}
\end{minipage}
\begin{minipage}[t]{0.19\linewidth}
%\centering
{
\centering
\includegraphics[width=3.4cm,height=7.8cm]{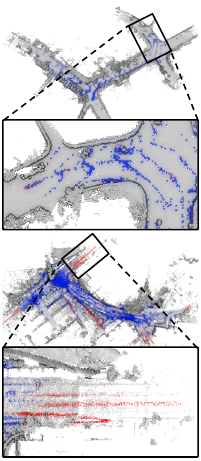}
\centerline{\small (b) DUFOMap\cite{10496850}}
}
\end{minipage}
\begin{minipage}[t]{0.19\linewidth}
%\centering
{
\centering
\includegraphics[width=3.4cm,height=7.8cm]{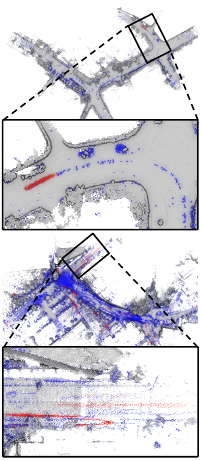}
\centerline{\small (c) Removert\cite{kim2020remove}}
}
\end{minipage}
\begin{minipage}[t]{0.19\linewidth}
%\centering
{
\centering
\includegraphics[width=3.4cm,height=7.8cm]{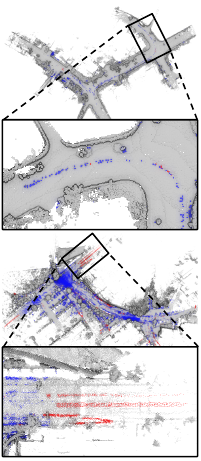}
\centerline{\small (d) ERASOR\cite{lim2021erasor}}
}
\end{minipage}
\begin{minipage}[t]{0.19\linewidth}
%\centering
{
\centering
\includegraphics[width=3.4cm,height=7.8cm]{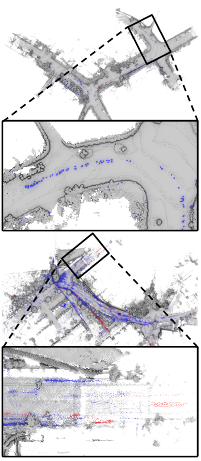}
\centerline{\small (e) FreeDOM (ours)}
}
\end{minipage}
\vspace{-0.4cm}
\caption{Comparison of the results conducted by our proposed method and state-of-the-arts on SemanticKITTI 02 and HeLiMOS Ouster 8292-8901. Red points are residual dynamic points and blue points are falsely removed static points. The fewer colored points, the better. }
\label{compare}
\vspace{-0.4cm}
\end{figure*}

\textbf{Static map construction performance:}
We compare our method against Octomap \cite{hornung2013octomap}, DUFOMap \cite{10496850} (post-processing mode), Removert \cite{kim2020remove}, ERASOR \cite{lim2021erasor} , and BeautyMap \cite{10533672}, as shown in Table \ref{table:1} and Fig. \ref{compare}. 

In SemanticKITTI, the visibility-based method OctoMap achieves the highest RR in sequence 02, but performs poorly in sequence 07, primarily due to the presence of numerous dynamic points lacking background observations, which cannot be filtered using visibility cue. Similarly, DUFOMap and Removert exhibit the same limitation. Although our method also relies on visibility cues, the incorporation of raycast enhancement allows it to remove most dynamic points without background observations, thereby maintaining a relatively high RR. ERASOR achieves high RR but suffers from low PR, as its 2D R-POD descriptor fails to accurately represent complex terrains, resulting in FPs. Additionally, due to \textit{incidence angle ambiguity}, both OctoMap and Removert generate a large number of FPs, leading to low PR. By employing spatial conservativeness, our method effectively mitigates \textit{incidence angle ambiguity} and achieve the highest PR. 

\begin{table}[t]
\begin{center}
\caption{Quantitative comparison on all LiDARs of HeLiMOS.}
\label{table:2}
\tabcolsep=2.5mm
\begin{tabular}{m{1.5cm}<{\centering} m{2.0cm}<{\centering} m{0.8cm}<{\centering} m{0.8cm}<{\centering} m{1.0cm}<{\centering}}
\toprule
LiDAR & \makecell[l]{Method} & PR(\%) & RR(\%) & F$_1$ score \\
\hline \multirow{6}{*}{\makecell{Ouster\\(dense)}}
    &\makecell[l]{OctoMap}          &83.60             &94.76             &88.82 \\
    &\makecell[l]{DUFOMap}          &92.28             &93.42             &92.85 \\
    &\makecell[l]{Removert}         &82.65             &92.03             &87.09 \\
    &\makecell[l]{ERASOR}           &90.37             &92.99             &91.66 \\
    &\makecell[l]{BeautyMap}        &94.48             &86.76             &90.45 \\
    &\makecell[l]{FreeDOM (Ours)}   &\textbf{98.28}    &\textbf{96.05}    &\textbf{97.15} \\
\hline \multirow{6}{*}{\makecell{Velodyne\\(sparse)}}
    &\makecell[l]{OctoMap}          &82.45             &82.46             &82.46 \\
    &\makecell[l]{DUFOMap}          &95.74             &68.27             &79.71 \\
    &\makecell[l]{Removert}         &88.01             &75.95             &81.54 \\
    &\makecell[l]{ERASOR}           &84.68             &76.25             &80.24 \\
    &\makecell[l]{BeautyMap}        &91.84             &81.42             &86.31 \\
    &\makecell[l]{FreeDOM (Ours)}   &\textbf{98.50}    &\textbf{83.97}    &\textbf{90.66} \\
\hline \multirow{6}{*}{\makecell{\hspace{-0.1cm}Livox\\\hspace{-0.1cm}(Non-repetitive)}}
    &\makecell[l]{OctoMap}          &92.42             &79.48             &85.46 \\
    &\makecell[l]{DUFOMap}          &96.26             &82.90             &89.08 \\
    &\makecell[l]{Removert}         &89.63             &73.55             &80.79 \\
    &\makecell[l]{ERASOR}           &89.23             &90.35             &89.79 \\
    &\makecell[l]{BeautyMap}        &84.79             &89.61             &87.13 \\
    &\makecell[l]{FreeDOM (Ours)}   &\textbf{97.85}    &\textbf{92.78}    &\textbf{95.25} \\
\hline \multirow{6}{*}{\makecell{Aeva\\(Narrow FoV)}}
    &\makecell[l]{OctoMap}          &82.89             &85.40             &84.13 \\
    &\makecell[l]{DUFOMap}          &91.40             &83.12             &87.07 \\
    &\makecell[l]{Removert}         &90.73             &68.32             &77.95 \\
    &\makecell[l]{ERASOR}           &84.69             &89.65             &87.10 \\
    &\makecell[l]{BeautyMap}        &85.38             &86.26             &85.82 \\
    &\makecell[l]{FreeDOM (Ours)}   &\textbf{97.34}    &\textbf{90.02}    &\textbf{93.54} \\
\bottomrule
\end{tabular}
\end{center}
\vspace{-0.8cm}
\end{table}

In HeLiMOS Ouster, the RR of OctoMap, DUFOMap, and Removert exhibit a noticeable decline compared to their performance in SemanticKITTI sequence 02, due to relatively fewer background observations in highly dynamic environments. The RR of ERASOR and BeautyMap also decreases. We believe that this is due to the increased environmental complexity, which cannot be accurately described using 2D descriptors. Benefiting from free space estimation, our method can quickly remove dynamic points by leveraging free space information before their arrival, and further eliminate residual dynamic points leveraging information after their arrival. Consequently, our method achieves the highest RR while maintaining the highest PR and F1 score. 

In Corridor, due to the reduced ratio of ground points, the performance of ERASOR and BeautyMap shows a noticeable decline. In Stairs, their performance declines significantly, as 2D descriptors fail to represent the 3D multi-layered environment. In contrast, visibility-based methods maintain consistent performance across indoor and outdoor settings. Our method overcomes the challenges of unstructured environments, achieving the highest PR, RR, and F1 score in Corridor, and the highest PR and F1 score in Stairs. 

We also conduct experiments on all LiDARs of HeLiMOS, which feature diverse characteristics, as shown in Table \ref{table:3}. All methods exhibit performance degradation compared to their performance on Ouster. Among them, a significant decline in RR is evident on Velodyne, primarily due to the sparsity of the point cloud and FoV occlusions, which result in insufficient environmental observations. Despite these challenges, our method achieves the highest PR, RR, and F1 score across all LiDARs, demonstrating its robustness.

\textbf{Ablation study:}
Ablation experiments are performed to verify the effectiveness of the raycast enhancement and map-refinement back-end. As shown in Table \ref{table:4}, incorporating the map-refinement back-end significantly improves the RR, resulting in an overall superior performance. Compared to the method without raycast enhancement, our method achieves a noticeable increase in RR on SemanticKITTI, particularly in sequence 07, as shown in Fig. \ref{raycast_enh}, which contains scenarios with limited background observations. Since such cases rarely occur in HeLiMOS, raycast enhancement does not result in notable performance changes on Ouster and Livox. Interestingly, on Velodyne, our method with raycast enhancement shows an increase in RR. We hypothesize that this is because the raycast enhancement partially compensates the sparsity of the sensor. Additionally, since local context is often sufficient for robotics applications, we also perform evaluations within a maximum range of 20 meters. The results show a significant improvement in RR across all sensors, particularly on Velodyne. This demonstrates that our method performs robustly in robotics applications, even on the sparsest LiDAR.

\begin{table}[t]
\begin{center}
\caption{Ablation study on KITTI and HeLiMOS.}
\label{table:3}
\tabcolsep=2mm
\begin{tabular}{m{1.8cm}<{\centering} m{2.2cm}<{\centering} m{0.8cm}<{\centering} m{0.8cm}<{\centering} m{1.0cm}<{\centering}}
\toprule
Dataset & \makecell[l]{Method} & PR(\%) & RR(\%) & F$_1$ score \\
\hline \multirow{4}{*}{KITTI}
    &\makecell[l]{w/o Raycast Enh.}    &99.36             &94.38             &96.81 \\
    &\makecell[l]{w/o Back-end}        &\textbf{99.54}    &77.94             &87.42 \\
    &\makecell[l]{FreeDOM (ours)}      &99.21             &\textbf{97.89}    &\textbf{98.55} \\ \cline{2-5}
    &\makecell[l]{FreeDOM (20m)}       &98.68             &\textbf{99.75}    &\textbf{99.21} \\
\hline \multirow{4}{*}{\makecell{HeLiMOS\\Ouster}}
    &\makecell[l]{w/o Raycast Enh.}    &98.25             &96.03             &97.13 \\
    &\makecell[l]{w/o Back-end}        &\textbf{98.97}    &79.10             &87.93 \\
    &\makecell[l]{FreeDOM (ours)}      &98.28             &\textbf{96.05}    &\textbf{97.15} \\ \cline{2-5}
    &\makecell[l]{FreeDOM (20m)}       &97.82             &\textbf{98.48}    &\textbf{98.15} \\
\hline \multirow{4}{*}{\makecell{HeLiMOS\\Velodyne}}
    &\makecell[l]{w/o Raycast Enh.}    &98.88             &82.63             &90.03 \\
    &\makecell[l]{w/o Back-end}        &\textbf{99.45}    &60.68             &75.37 \\
    &\makecell[l]{FreeDOM (ours)}      &98.50             &\textbf{83.97}    &\textbf{90.66} \\ \cline{2-5}
    &\makecell[l]{FreeDOM (20m)}       &97.39             &\textbf{91.00}    &\textbf{94.09} \\
\hline \multirow{4}{*}{\makecell{HeLiMOS\\Livox}}
    &\makecell[l]{w/o Raycast Enh.}    &97.80             &92.76             &95.21 \\
    &\makecell[l]{w/o Back-end}        &\textbf{98.46}    &68.68             &80.92 \\
    &\makecell[l]{FreeDOM (ours)}      &97.86             &\textbf{92.78}    &\textbf{95.25} \\ \cline{2-5}
    &\makecell[l]{FreeDOM (20m)}       &97.58             &\textbf{94.68}    &\textbf{96.11} \\
\hline \multirow{4}{*}{\makecell{HeLiMOS\\Aeva}}
    &\makecell[l]{w/o Raycast Enh.}    &97.39             &89.91             &93.50 \\
    &\makecell[l]{w/o Back-end}        &\textbf{98.62}    &62.60             &76.59 \\
    &\makecell[l]{FreeDOM (ours)}      &97.34             &\textbf{90.02}    &\textbf{93.54} \\ \cline{2-5}
    &\makecell[l]{FreeDOM (20m)}       &97.12             &\textbf{93.51}    &\textbf{95.28} \\
\bottomrule
\end{tabular}
\end{center}
\vspace{-0.3cm}
\end{table}

\begin{table}[t]
\begin{center}
\caption{Real-time performance on all LiDARs.}
\label{table:4}
\tabcolsep=2mm
\begin{tabular}{m{1.2cm}<{\centering} m{2.5cm}<{\centering} m{1.8cm}<{\centering} m{1.8cm}<{\centering}}   
\toprule
\makecell[l]{Dataset} & \makecell[l]{LiDAR} & Runtime (ms) & Frequency (hz) \\   
\hline   
\makecell[l]{KITTI}       &\makecell[l]{Velodyne HDL-64}      &62.042     &16.118 \\
\makecell[l]{HeLiMOS}     &\makecell[l]{Ouster OS2-128}       &90.335     &11.070 \\
\makecell[l]{HeLiMOS}     &\makecell[l]{Velodyne VPL-16C}     &26.270     &38.067 \\
\makecell[l]{HeLiMOS}     &\makecell[l]{Livox Avia}           &17.883     &55.920 \\
\makecell[l]{HeLiMOS}     &\makecell[l]{Aeva Aeries II}       &19.132     &52.268 \\
\makecell[l]{Corridor}    &\makecell[l]{Robosense Helios-32}  &19.778     &50.561 \\
\makecell[l]{Stairs}      &\makecell[l]{Livox mid-360}        &15.691     &63.731 \\
\bottomrule
\end{tabular}   
\end{center}
\vspace{-0.8cm}
\end{table}

\textbf{Algorithm speed:}
As real-time processing is necessary for autonomous robotics applications, we measure the average processing time on each dataset using an Intel i9-13900HX laptop CPU, as shown in Table \ref{table:4}. Among all datasets, the HeLiMOS Ouster dataset features the largest scenes and the highest point density, resulting in the longest processing time. Despite these challenges, our method maintains an average processing frequency exceeding 10 Hz, meeting the typical LiDAR frame rate for real-time applications. 

\section{Conclusions}
In this paper, we propose FreeDOM, a novel online dynamic object removal framework based on conservative free space estimation. As evaluated across diverse scenarios and sensors, our method overcomes the limitations of visibility-based methods and achieves state-of-the-art performance. 

\section*{Acknowledgment}
We would like to thank Xiangyu Cheng, Jingchao Feng, Yifan Feng, Letian Fu, Yuhao Liu, Anjia Wang, and Xianwei Yuan for their assistance with data collection. 

\bibliographystyle{IEEEtran}
\bibliography{IEEEabrv,ref}
\end{document}